\documentclass{article} 
\usepackage{nips15submit_e,times}
\usepackage{hyperref}
\usepackage{url}
\usepackage{graphicx}
\usepackage{amsmath}

\title{Structured Memory for Neural Turing Machines}

\author{
Wei Zhang 
\thanks{Alternative email for Wei Zhang: wynnzh@gmail.com} 
\quad Yang Yu\quad Bowen Zhou \\
IBM Watson\\
\texttt{zhangwei,yu,zhou@us.ibm.com} \\
}

\nipsfinalcopy 

\begin{document}

\maketitle

\begin{abstract}
Neural Turing Machines (NTM) [2] contain memory component that simulates ``working memroy'' in the brain to store and retrieve information to ease simple algorithms learning. So far, only linearly organized memory is proposed, and during experiments, we observed that the model does not always converge, and overfits easily when handling certain tasks. We think memory component is key to some faulty behaviors of NTM, and better organization of memory component  could help fight those problems. In this paper, we propose several different structures of memory for NTM, and we proved in experiments that two of our proposed structured-memory NTMs could lead to better convergence, in term of speed and prediction accuracy on copy task and associative recall task as in [2].
\end{abstract}

\section{Introduction}
Memory components for Neural Networks has been recently introduced in Neural Turing Machines (NTMs) [2], Memory Networks [1], and Dynamic Memory Networks [3]. The purposes of those memory components are similar, which is to simulate ``working memory'' in the brain to store temporary information through time to be used by attention module for reading and writing. In this paper, our focus is on NTMs. They are attractive in that memory could be randomly accessed by controller through blurry ``erase'' and ``write'' operations, which well aligns with Turing Machine operations. When memory size is small, nice convergence of the NTM model could be  frequently observed, although not always. But when memory size is large, the model struggles to convergence, and sometimes the test loss jumps drastically in a large range, which is a sign of overfitting, as was observed in our experiments. We think that the dynamics of memory contents takes a crucial role in controlling model convergence speed and quality. Thus, we proposed and experimented different memory structures to explore if specific memory structure could lead to more stable memory contents, or say, perform ``memory smoothing'' so that the memory content generated after read and write operations does not deviate too far from the ``expected'' memory content, so that overfitting of parameters could be in turn alleviated. We proposed three different architectures, NTM1, NTM2 and NTM3, which are explained in detail in section \ref{sec:arch}, and experiments are shown in section \ref{sec:exp}.

\begin{figure}
\begin{center}
\includegraphics[scale=0.55]{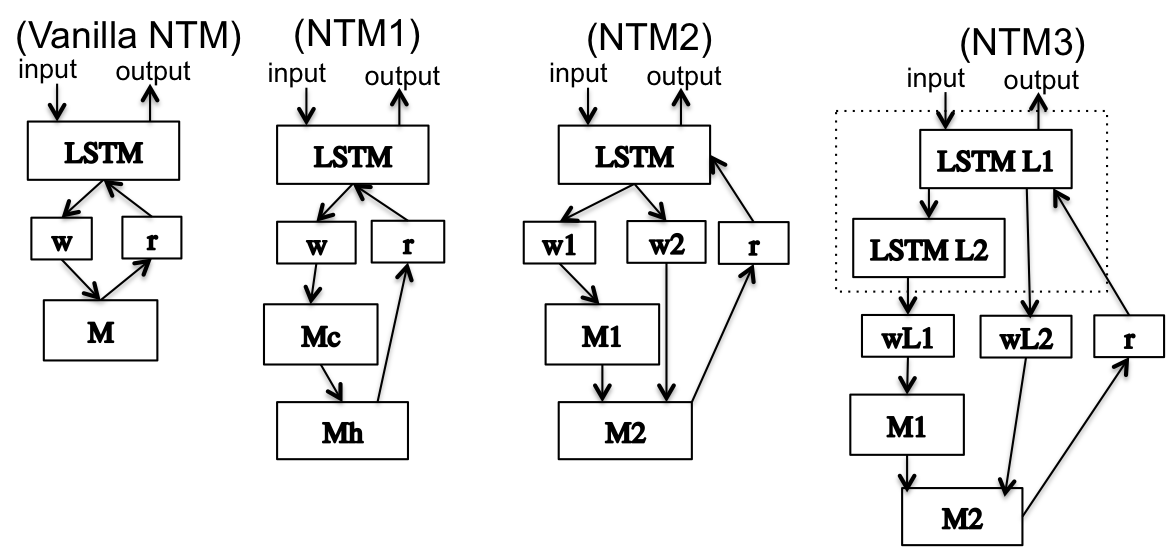}
\caption{NTM and NTM variants that use Long short-term memory [5] as controllers. Note that every module in those modules are updated recurrently through time using their previous states. NTM1 contains only one write head, and one of the memory $\mathbf{M}_h$ is not written directly by controller, which is different from $\mathbf{M}_2$ in NTM2 that is written both by $\mathbf{M}_1$ and write head simultaneously. NTM3 is different from NTM2 in that NTM3 write heads takes inputs from two layers. NTM3 could also be expanded to multiple layers as well.}
\label{fig}
\end{center}
\end{figure}

\section{Architectures}\label{sec:arch}
We show the NTM original architecture introduced by Graves et al. [2] and three variants NTM1, NTM2 and NTM3 in Figure \ref{fig}. Details of each model is explained in a moment. We introduce a notion of \textit{Memory visibility}: We call memory ``Controlled'' if it is modified by controller outputs through write heads directly, or ``Hidden'' if not. 

\textbf{NTM1} Compared to NTM, NTM1 has an additional hidden memory $\mathbf{M}_h$, which is not controlled by controller module, but is connected to the controlled memory $\mathbf{M}_c$. The hidden memory accumulates the content in the controlled memory, so that a type of memory smoothing is performed to prevent memory from deviating from the ``expected content''. Specifically, the memory content for hidden memory $\mathbf{M}_h(t)$ of time $t$ is generated as:
\begin{align*}
write: \quad &\mathbf{M}_c(t)= h \big( \mathbf{M}_c(t-1), \mathbf{w}(t-1), \mathbf{c}(t) \big)  \\
update: \quad &\mathbf{M}_h(t) =a \mathbf{M}_h(t-1) + b \mathbf{M}_c(t) \\
read: \quad  &\mathbf{r}(t) = \mathbf{w_r}(t) \mathbf{M}_h(t)
\end{align*}
$\mathbf{M}_c$ is updated by $t-1$ time write head to generate $t$ time controlled memory output, which in turn is used to update hidden memory $\mathbf{M}_h$, and then the new hidden memory is used to generate read head at time $t$.  $\textbf{w}(t)$ is write weights, $\textbf{c}(t)$ is the controller output of time $t$, and $h()$ is the function that updates controlled memory and write weights that implement ``erase and add'' operations. $\mathbf{w_r}$ is read weights. $a$ and $b$ are scalar mixture weights, which could be further extended to tensors. We use scalars in this work. The read head is reading from $\mathbf{M}_h(t)$ instead of $\mathbf{M}(t)$ as is did in [2].

\textbf{NTM2} The second architecture NTM2 is similar to NTM1 in that two memory blocks are used, and they are connected hierarchically. However, the difference from NTM1 is that hidden memory in NTM2 is no longer hidden from the controller, but connected to controller outputs through another write head. So two memory blocks are all controlled memories connecting to two different write heads. The upper level memory is denoted as $\mathbf{M}_1$, and the lower level one as $\mathbf{M}_2$. $\mathbf{M}_1$ is modified solely by L1 write head, but $\mathbf{M}_2$ is modified by both $\mathbf{M}_{1}$ contents and L2 write head. The logic is:
\begin{align*}
write:\quad  &\mathbf{M}_{1}(t), \mathbf{w}_1(t) = h \big( \mathbf{M}_{1}(t-1), \mathbf{w}_1(t-1), \mathbf{c}(t) \big) \\
&\widetilde\mathbf{M}_{2}(t), \mathbf{w}_2(t) = h \big( \mathbf{M}_{2}(t-1), \mathbf{w}_2(t-1), \mathbf{c}(t) \big)  \\
update:\quad  &\mathbf{M}_{2}(t) =a \widetilde\mathbf{M}_{2}(t) + b \mathbf{M}_{1}(t)\\
read:\quad  &\mathbf{r}(t) = \mathbf{w_r}(t) \mathbf{M}_{2}(t)
\end{align*}

Additionally, $\mathbf{M}_{1}$ and $\mathbf{M}_{2}$ are all generated by the same head function with same controller input, but with different write weights. Single read head $\mathbf{r}$ is used, which means the output is only read from the lowest layer memory ($\mathbf{M}_{2}$ in this case). Note that the architecture could be easily expanded into multiple layers by increasing number of heads and memory blocks.

\textbf{NTM3} The third architecture is significantly different from the previous two. First, the controller in the model have multiple layers. Second, each layer output is connecting to a memory by write heads. Different layers of memory contains different level of transformation of the input. In the case where multi-layer LSTM is used as NTM controller, output from each layer goes through the non-linear transformation of write heads, and writes to each individual memory. Then the deeper level memory is updated by upper level memory and write head jointly. 
\begin{align*}
write:\quad &\mathbf{M}_{1}(t), \mathbf{w}_{L1}(t) = h \big( \mathbf{M}_{1}(t-1), \mathbf{w}_{L1}(t-1), \mathbf{c}_{L1}(t) \big)  \\
&\widetilde\mathbf{M}_{2}(t), \mathbf{w}_{L2}(t) = h \big( \mathbf{M}_{2}(t-1), \mathbf{w}_{L2}(t-1), \mathbf{c}_{L2}(t) \big)  \\
update:\quad & \mathbf{M}_{2}(t) =  a \widetilde{\mathbf{M}}_{2}(t) + b {\mathbf{M}}_{1}(t)\\
read:\quad & \mathbf{r}(t) = \mathbf{w_r}(t) \mathbf{M}_{2}(t)
\end{align*}

 In so doing, the memory blocks receives write operations not only from the final layer LSTM output, but also from the intermediate layer outputs as well. The purpose is to smooth the final layer memoroy with intermediate LSTM outputs. We can see in a moment how those three performs on copy and associative recall task.

\begin{figure}
\begin{center}
\includegraphics[scale=0.3]{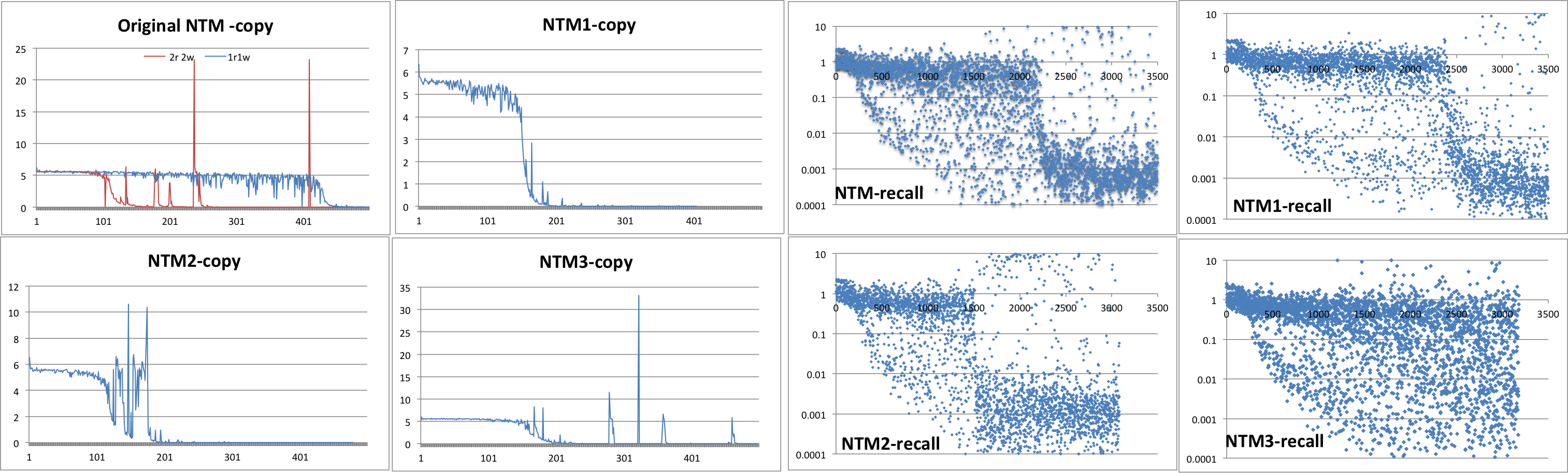}
\caption{comparison of convergence speed and quality on copy task and associative recall task. Four figures to the left are all from copy task, and four on the right are for associative recall task. For all the figures, X-axis is the number of iterations (scaled by sampling once for every 25 iterations), and for copy task, Y-axis shows the binary cross entropy loss per item (sequence of length 8), and in recall task, Y-axis shows log of binary cross entropy loss. Memory is 128 vectors of length 20, for every memory block shown in Figure \ref{fig}.}
\label{fig:converge}
\end{center}
\end{figure}

\section{Experiments}\label{sec:exp}
The experiments is to show the convergence speed and quality of those three variants, compared to the NTM setting. In theory, the convergence should happen for every run of every model, which is not the case in practice. Randomness of parameter initialization might be the reason\footnote{We applied the technique used by https://github.com/kaishengtai/torch-ntm to initialize each memory slot to define ``write/read order'' prior}. But if model convergences, the number of iterations used were quite stable across runs. Thus, to make the evaluation reliable, we repeat experiments at least 5 times for each model and choose the most frequent circumstance for showing.

The tasks we choose for model testing are copy task and associative recall task as in [2]. Associative recall task requires the model to target an item in sequence, and later on shift to the immediate next item to give as an answer. Copy task requires the model to consecutively output items in original input order, which requires long-term location based addressing capability. In each training iteration, we randomly generated each item as a binary vector of length 8.  The number of items is random as well. In recall task, the items, and number of items to be generated are all chosen at random. All those randomness is to guarantee that every training iteration uses different training example, so that overfit to specific data set is no longer an issue, to guarantee that algorithms could be learned. This is critical to our further analysis.For all the models, the training criteria is binary cross entropy, and we use RMSProp [4] for optimization, with learning rate $1E-4$, momentum $0.9$, and decay $0.95$.

In Figure \ref{fig:converge} we show our results on copy and recall respectively. Copy tasks are four graphs to the left. X-axis shows the number of iteration used (sampled every 25 iterations), and Y-axis is the binary cross entropy loss of each iteration. The shown result in each graph is representative among 5 runs, among which the selected curve is observed most frequently. We can see that NTM1 and NTM2 shows faster convergence and less outliers above the convergence range (closely around 0 near X-axis) than NTM or NTM3. The outliers (extremely high loss) means that the prediction on the item in that iteration is significantly incorrect, which is a sign of overfitting or underfitting. In NTM1 and NTM2 runs we see very few outliers, but NTM shows more outliers when using two read and write heads (the curve with sharp values), or does not converge as fast when using 1 read and 1 write head (flat curve). This pattern is consistently observed over 5 runs as well. We can also read out from Figure ``NTM-copy'' that using 1 read and 1 write head converges significantly slower than 2w/r heads. The series of experiments on copy task confirms our assumption that the introduction of additional memory in NTM1 and NTM2 do help stabilize the memory component, which in turn leads to better tuning of the parameters. However, NTM3 does not show nice curves, and produce outlier almost every 1000 iterations or so. This means that the memory that is attached to the intermediate layers of LSTM introduces more noise, which is unexpected. Although less stable, NTM does converge about 500 iterations faster than NTM1 and NTM2, but we observed the opposite in associative recall task.

For associative recall, we can see that outliers are produced much more frequently when loss significantly reduces, and we rarely observe convergence of original NTM or NTM3. In ``NTM3-recall'' graph, we show a non-converged run since it happens frequently for NTM3. But NTM1 and NTM2 converges more frequently across runs. And NTM2 shows much faster convergence, roughly with 37,000 iterations, compared to 50,000 for NTM or NTM1. Moreover, NTM1 and NTM2 generate less outliers than NTM. This means that, the prediction accuracy for NTM1 and NTM2 will be higher than NTM, since more training examples are correctly predicted than NTM. And, NTM1 and NTM2 will have higher probability to be a non-overfitting model than NTM. Unfortunately, NTM3 model does not converge as frequent as NTM1 and NTM2, which aligns with the observation in copy task about the difference of those models. 

\section{Conclusion}
This paper discussed three new structured memory architectures for Neural Turing Machines, and showed that organizing memory blocks in a proper hierarchical manner could alleviate overfitting and sometimes increase predictive accuracy compared to NTM. 

\section{Future work}
In the future we would also try NTM1 NTM2 on data sets that require more complex reasoning. We tested NTM on a synthetic QA data set proposed in [1], and observed the instability in convergence. We would try NTM1 and NTM2 on the same data set.
\section*{References}

[1] Jason Weston, Sumit Chopra, \& Antoine Bordes. Memory Networks. \textit{International Conference on Representation Learning}, 2015

[2] Alex Graves, Greg Wayne, \& Ivo Danihelka. Neural Turing Machines. \textit{arXiv preprint arXiv:1410.5401.} 2014

[3] Ankit Kumar,  Ozan Irsoy, Jonathan Su, James Bradbury, Robert English, Brian Pierce, Peter Ondruska, Ishaan Gulrajani, \& Richard Socher. Ask Me Anything: Dynamic Memory Networks for Natural Language Processing. \textit{arXiv preprint arXiv:1506.07285.} 2015

[4] Tijmen Tieleman, \& Geoffrey Hinton. "Lecture 6.5-rmsprop: Divide the gradient by a running average of its recent magnitude." \textit{COURSERA: Neural Networks for Machine Learning} 4. 2012 

[5] Sepp Hochreiter, \& Jürgen Schmidhuber. Long short-term memory. \textit{Neural computation}, no. 8: 1735-1780. 1997
\end{document}